\documentclass[letterpaper, 10 pt, conference]{ieeeconf}  

\usepackage{amsmath}
\usepackage{amssymb}
\usepackage{moreverb,url}
\usepackage[]{hyperref}
\usepackage{algorithm}
\usepackage{algpseudocode}
\usepackage{dirtytalk}
\usepackage{graphicx} 
\usepackage{subcaption}
\usepackage{array}
\usepackage{tabularx}
\usepackage{url}
\usepackage{booktabs}

\IEEEoverridecommandlockouts                              

\title{\LARGE \bf Learning What Matters Now: A Dual‑Critic Context‑Aware RL Framework for Priority‑Driven Information Gain}

\author{Dimitris Panagopoulos$^{1}$, Adolfo Perrusqu\'{i}a$^{1}$ and Weisi Guo$^{1}$
\thanks{$^{1}$Faculty of Engineering and Applied Science, Cranfield University, Cranfield MK43 0AL, UK,
        {\tt\small \{d.panagopoulos, adolfo.perrusquia-guzman, weisi.guo\}@cranfield.ac.uk}. 
        This work is funded by EPSRC iCASE with Thales UK (EP/X52475X/1)}%
}

\begin{document}
\setlength{\parskip}{0em}

\maketitle
\thispagestyle{empty}
\pagestyle{empty}

\begin{abstract}
Autonomous systems operating in high‑stakes search‑and‑rescue (SAR) missions must continuously gather mission‑critical information while flexibly adapting to shifting operational priorities. We propose CA‑MIQ (Context‑Aware Max‑Information Q‑learning), a lightweight dual‑critic reinforcement learning (RL) framework that dynamically adjusts its exploration strategy whenever mission priorities change. CA‑MIQ pairs a standard extrinsic critic for task reward with an intrinsic critic that fuses state‑novelty, information‑location awareness, and real‑time priority alignment. A built‑in shift detector triggers transient exploration boosts and selective critic resets, allowing the agent to re‑focus after a priority revision. In a simulated SAR grid‑world, where experiments specifically test adaptation to changes in the priority order of information types the agent is expected to focus on, CA‑MIQ achieves nearly four times higher mission‑success rates than baselines after a single priority shift and more than three times better performance in multiple‑shift scenarios, achieving 100\% recovery while baseline methods fail to adapt. These results highlight CA‑MIQ’s effectiveness in any discrete environment with piecewise‑stationary information‑value distributions.
\end{abstract}


\section{Introduction}
Search and rescue (SAR) operations in disaster scenarios present extraordinary challenges for autonomous agents. These environments are inherently dynamic, with conditions that evolve rapidly as new information becomes available. Traditional reinforcement learning (RL) methods for robotic exploration, whether $\varepsilon$-greedy, Boltzmann, or even maximum-entropy variants, often assume a stationary environment with fixed priorities, failing to account for the fluid nature of SAR missions where the relative importance of information types can shift abruptly as the situation evolves \cite{ladosz2022exploration, ewers2025deep}. For instance, the discovery of a gas leak may suddenly elevate ``hazard assessment B'' to higher priority than ``victim status A.'' Similarly, the identification of an unstable structure might necessitate immediate exploration of alternative access routes. Existing approaches to information-directed exploration, while effective in stationary environments, lack mechanisms for adapting to these priority shifts \cite{sukhija2024maxinforl, hao2022regret}. This results in agents that continue sampling from distributions no longer aligned with operational needs leading to wasted time, unnecessary risk exposure, and suboptimal triage outcomes.

Recent advances have begun addressing these challenges through different approaches. Panagopoulos et al. \cite{panagopoulos2024selective} combine a hierarchical decision-making architecture with RL and large language model (LLM)  that formalises information gathering task within a structured Information Space. While this formulation enables agents to collect information in a systematic manner, it assumes a static ordering of priorities throughout the mission. This limitation significantly restricts the practical utility of autonomous agents in real-world disaster scenarios where responders must continuously reassess and reprioritise objectives based on emerging information. In a complementary direction, the authors in \cite{sukhija2024maxinforl} demonstrate that exploration strategies maximising information gain significantly improve sample-efficiency in continuous control tasks by directing agents toward transitions with highest uncertainty. However, their framework is formulated for high-dimensional function-approximation settings and does not address mission-level priority shifts in the discrete, tabular domains that characterise high-level decision-making \cite{rahman2022adversar}.

These limitations reveal a critical gap. Systems guided by human knowledge can incorporate and exploit stakeholder bias but cannot effectively re-explore when information is revised, leading to rigid behavior that fails to adapt to changing conditions \cite{chen2025target, zhou2024multi}. While information-gain driven methods can trigger purposeful exploration, they lack mechanisms for aligning that exploration with dynamic, task-level priorities, resulting in agents that explore efficiently but not necessarily in directions aligned with current mission objectives. Bridging this gap requires an exploration policy that (1) recognises when its current ordering of information targets becomes obsolete, and (2) focuses exploration effort toward prioritised transitions without discarding still-relevant knowledge.

To address these limitations, we introduce CA‑MIQ (Context‑Aware Max‑Information Q‑learning), a framework that unifies information-gain maximisation with priority-aware adaptation in discrete state-action space environments. Our approach treats priority ordering as a latent context variable that can shift during operation, and implements three key technical advances. First, we extend the Information Space formulation \cite{panagopoulos2024selective} to model priority ordering as a latent context variable. The agent monitors performance statistics to detect context shifts or receives explicit priority updates from operators, enabling online re-focus on what matters right now. Second, CA-MIQ maintains two tabular critics: an \textit{extrinsic} that learns task reward as usual, and an \textit{intrinsic} that learns an information-gain surrogate that combines state-visit novelty, information-location awareness, and priority alignment. The latter measures how well an action reduces uncertainty about currently high-priority information. This architecture builds on MaxInfoRL \cite{sukhija2024maxinforl} but adapts it under different unknowns and at different scales to tackle a different point: \textit{how do you keep exploring the right facts when mission priorities keep shuffling?}. Instead of trying to reduce the global model uncertainty, which implies chasing epistemic uncertainty over the physics of the entire environment (continuous low-level control), we target the mission critical which is what matters most right now (discrete high-level control). Third, upon detecting a shift in priorities, our framework temporarily boosts exploration through a transient $\varepsilon$-increase that decays as the new ordering is learned, while partially resetting only the critic entries affected by the reordering.

Our experimental results demonstrate that CA-MIQ significantly outperforms traditional tabular methods while the resulting exploration policy maintains standard tabular learning convergence guarantees under piecewise-stationary priority schedules. In a simulated SAR environment, CA-MIQ reaches post-shift optimal performance almost four and three times faster than both flat baselines with and without exploration boosts in single and multiple priority shifts scenarios, respectively. By bridging the gap between static information structures and dynamic operational realities, this work represents an important step toward more adaptive autonomous systems capable of maintaining effectiveness in evolving environments. To the best of our knowledge, this work is the first to introduce an information‑gain–driven exploration strategy that responds to piecewise‑stationary shifts in mission‑level priorities within SAR environments.


\section{Related Work}
Traditional exploration methods in RL include $\varepsilon$-greedy \cite{sutton1998reinforcement}, Boltzmann exploration \cite{cesa2017boltzmann}, and upper confidence bound (UCB) approaches \cite{auer2002using}. While these methods provide theoretical guarantees for stationary environments, they struggle when faced with priority shifts or non-stationary reward structures. More sophisticated approaches like Thompson sampling \cite{chapelle2011empirical} offer probabilistic exploration but similarly assume fixed reward distributions.

Intrinsic motivation has emerged as a powerful framework for exploration, using curiosity \cite{pathak2017curiosity}, novelty \cite{bellemare2016unifying}, and surprise \cite{achiam2017surprise} as internal rewards. Notably, Burda et al. \cite{burda2018exploration} demonstrated that curiosity-driven exploration can effectively navigate environments with sparse rewards. However, these approaches generally lack mechanisms to align exploration with shifting operational priorities.

Adapting to changing contexts remains challenging in RL. Meta-reinforcement learning \cite{finn2017model} attempts to learn policies that can quickly adapt to new tasks, while contextual bandits \cite{lattimore2020bandit} explicitly model context variables to improve decision-making. Hidden parameter Markov Decision Processes (MDPs) \cite{doshi2016hidden} and contextual MDPs \cite{hallak2015contextual} formalise environments where underlying parameters may change, though they typically assume the context is fully observable or inferred from state transitions. Recent work \cite{ren2022reinforcement} investigates adaptation in non-stationary environments by explicitly modeling context switches. Similarly, Lee et al. \cite{lee2020context} developed a context-aware exploration method that adapts to changing reward structures, though primarily in continuous domains using deep RL.

Information-theoretic approaches provide principled methods for efficient exploration. Information-Directed Sampling (IDS) \cite{russo2018learning} balances expected regret against information gain about the optimal action. Maximum entropy RL \cite{haarnoja2018soft} encourages exploration by maximising policy entropy alongside expected return.


\begin{figure}[!t]
\centering
\includegraphics[width=\columnwidth]{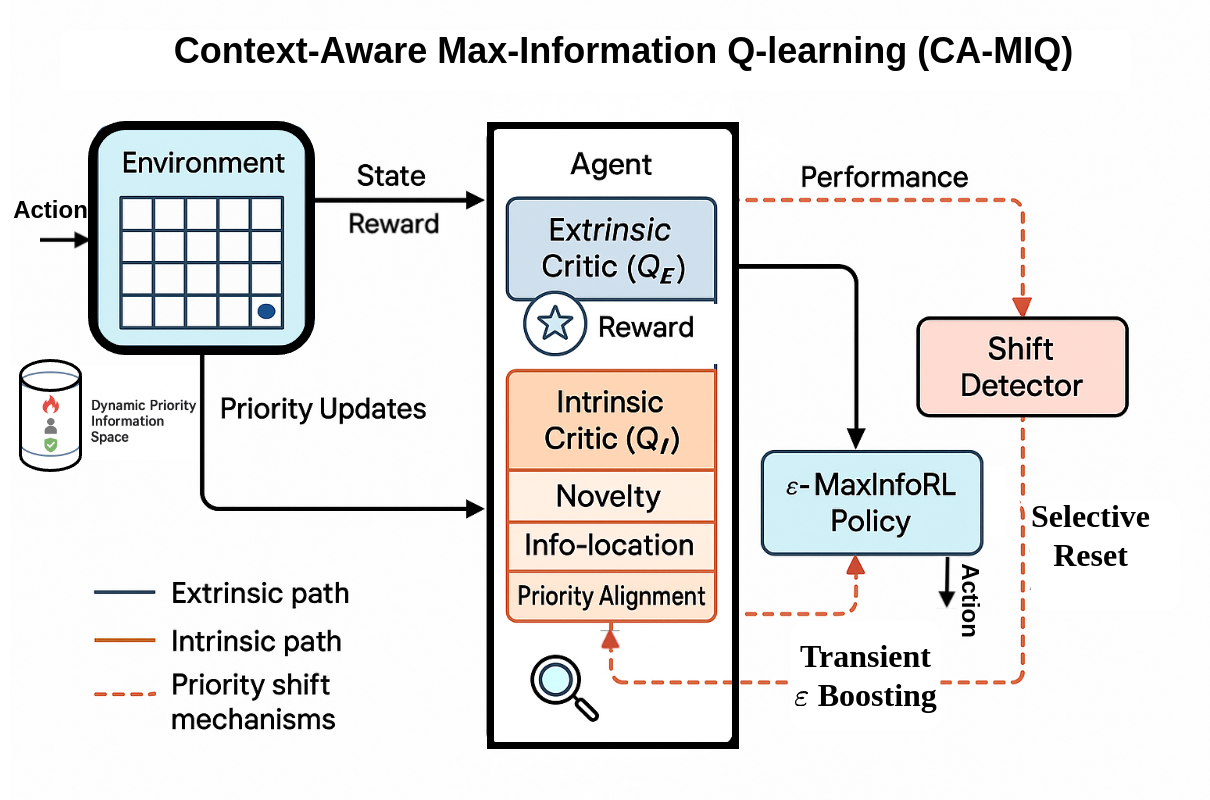}
\caption{\textbf{CA‑MIQ (Context‑Aware Max‑Information Q‑learning) architecture.} The environment supplies the current state to an agent equipped with two critics: an \textit{extrinsic critic} \(Q^{E}\) that evaluates task reward-seeking behaviour (black path) and an \textit{intrinsic critic} \(Q^{I}\) that aggregates novelty, information-location and priority-alignment signals (orange path). An \( \varepsilon\)-MaxInfoRL policy gate arbitrates between the critics to choose each action. A \textit{shift detector} monitors performance; when a distributional change is detected it triggers a dashed orange \textit{priority-shift loop} that transiently boosts exploration (\(\varepsilon\)-boosting) and \textit{resets} selectively the critics. Solid arrows denote data flow, whereas dashed arrows denote adaptive control signals.}
\label{fig1:framework}
\end{figure}

\section{Methodology}\label{sec:methodology}
\subsection{Problem Formulation}
We consider a tabular RL setting in a SAR scenario, modeled as an MDP \(\mathcal{M} = (S, A, P, R, \gamma)\), where \(S\) is the state space, \(A\) is the action space, \(P:S \times A \times S \rightarrow[0, 1]\) is the transition function, \(R: S \times A \rightarrow \mathbb{R}\) is the reward function, and \(\gamma \in [0 ,1)\) is the discount factor. Unlike standard MDPs, we explicitly model the collection of different types of information as a critical component of the agent's task. Similalry to \cite{panagopoulos2024selective}, this is formalised through an Information Space \(\mathcal{I} = \{i_1, i_2, \dots, i_n\}\), where each \(i_k\) represents a specific type of information (e.g., victim location, hazard assessment, safe zone) that the agent can collect. Crucially, each information type has an associated priority \(p_k \in \mathbb{R^{+}}\) that determines its relative importance. The collection priority ordering is initially defined as a mapping \(\mathcal{O}_t: \mathcal{I} \rightarrow \{1, 2, \dots, |\mathcal{I}|\}\), where \(\mathcal{O}_t(i_k)\) indicates the position of information type \(i_k\) in the collection sequence. Each item \(i_k \in \mathcal{I}\) resides at a location and is labelled with the current order \(\mathcal{O}_t(i_k)\). This ordering may change during operation, formally when \(\mathcal{O}_t \neq \mathcal{O}_{t-1}\), reflecting the evolving nature of SAR missions.

\subsection{Dynamic-Priority Information Space}
A key innovation in our approach is the modeling of the priority ordering as a latent context variable that can shift during operation. The information collection task is the triplet \((\mathcal{I}, \mathcal{O}, \mathcal{C})\), where \(\mathcal{I}\) and \(\mathcal{O}\) are the set of information types and the current priority ordering as defined above, respectively; \(\mathcal{C}\) is the finite set of feasible priority orderings (contexts). The current context determines which collection order the agent should follow. Context shifts occur when priorities change due to new information from human operators or environmental changes, defined as an update \(\Delta\pi\) at arbitrary time-steps. We model the MDP as conditionally stationary given a fixed context, with piecewise-stationary dynamics overall. The agent's objective is to maximise total reward (mission success) while minimising the adaption time to changes in the context \(\mathcal{O}\) after each shift.

\subsection{Dual-Critic Intrinsic-Extrinsic Architecture}\label{subsec:camiq}
To achieve adaptive information-directed exploration, we borrow the insight of \cite{sukhija2024maxinforl} so that the agent keeps two tabular critics:

\begin{enumerate}
    \item \textbf{Extrinsic critic} \textit{\(Q^E(s,a)\)}: Estimates the expected return from task rewards.
    \item \textbf{Intrinsic critic} \textit{\(Q^I(s,a)\)}: Estimates information gain about the MDP, favoring actions that reduce dynamic, priority-driven uncertainty.
\end{enumerate}

The full update equations for the tabular critics are as follows:

\begin{equation}\label{eq:original_q_update_q_extrinsic}
\begin{split}
Q^{E}(s_t, a_t) & \leftarrow Q^{E}(s_t, a_t) + \alpha [r_{t} + \\
& \gamma\max_{a'} Q^{E}(s_{t+1}, a') - Q^{E}(s_t, a_t)]  
\end{split}
\end{equation}

\begin{equation}\label{eq:original_q_update_q_intrinsic}
\begin{split}
Q^{I}(s_t, a_t) & \leftarrow Q^{I}(s_t, a_t) + \alpha [r_{int}(s_t, a_t) + \\
& \gamma\max_{a'} Q^{I}(s_{t+1}, a') - Q^{I}(s_t, a_t)]  
\end{split}
\end{equation}

Where \(r_{int}(s_t, a_t)\) is the intrinsic reward, computed as a weighted combination of three distinct components:

\begin{itemize}
    \item \textbf{\textit{Novelty-based exploration}}: Incentivises visiting rarely tried state–action pairs, preventing premature exploitation and seeding knowledge of alternative routes.

    \begin{equation}
        r_{\text{novelty}}(s,a) = \frac{\beta_1}{\sqrt{N(s,a)}}
    \end{equation}
    where \(N(s,a)\) is the visit count for the state-action pair.

    \item \textbf{\textit{Information location awareness}}: Steers exploration toward regions where mission‑critical information items are placed, so the robot does not waste effort in redundant and costly exploration.

    \begin{equation}
        r_{\text{info}}(s,a) =
            \begin{cases}
            \frac{\beta_2}{\sqrt{N_{\text{info}}(s)}}, & \text{if } s \text{ is an info location} \\
            0, & \text{otherwise}
            \end{cases}
    \end{equation}
    where $N_{\text{info}}(s)$ tracks visits specifically to information locations.

    \item \textbf{\textit{Priority alignment}}: Acts like a compass: it boosts actions that respect the live priority sequence and discourages picking up lower priority facts too early.

    \begin{equation}
        r_{\text{alignment}}(s,a) =
            \begin{cases}
            \beta_3, & \text{if ordered collection} \\
            -\beta_4, & \text{if out-of-order collection} \\
            0, & \text{otherwise}
            \end{cases}
    \end{equation}
\end{itemize}

where \(\beta_1, \beta_2, \beta_3, \beta_4 \in \mathbb{R}\) are tunable parameters.
The agent's action selection is performed using an \(\varepsilon\)-MaxInfoRL policy:

\begin{equation}
    a_t =
        \begin{cases}
        \arg\max_a Q^{I}(s_t, a) & \text{with probability } \varepsilon_t \\
        \arg\max_a Q^{E}(s_t, a) & \text{else,}
        \end{cases}
\end{equation}

Exploration therefore favours actions that promise high information gain and respect the present priority order. This approach differs fundamentally from standard \(\varepsilon\)-greedy exploration. While traditional exploration chooses random actions, our intrinsic critic \(Q^I\) guides exploration toward state-action pairs that maximise information gain, with specific emphasis on novel state-actions, information-rich locations, and actions that align with the current collection priorities.

\subsection{Priority-Shift Adaptation Mechanism}

\subsubsection{Shift Detection}
The agent monitors its collection success rate and maintains historical performance metrics. A context shift is detected either through explicit operator input or when the agent observes a significant drop in collection success alignment.

\subsubsection{Transient Epsilon Boosting}
Upon detecting a shift, the agent temporarily increases the exploration parameter \(\varepsilon\) by a factor of \(\lambda_{\text{boost}}\):
\begin{equation}
    \varepsilon_t \leftarrow \min(\varepsilon_{\text{MAX}}, \varepsilon_t \cdot \lambda_{\text{boost}})
\end{equation}
This adjustment increases the likelihood of selecting the exploratory branch of the \(\varepsilon\)-MaxInfoRL policy, which relies on the intrinsic critic \(Q^I\). As a result, the agent is more likely to pursue actions driven by information gain to adapt to new priorities. The boosted exploration decays gradually over \(D_{\text{boost}}\) episodes:
\begin{equation}\label{eq:epsilon}
    \varepsilon_{t+k} = \varepsilon_t \cdot \exp(-k/D_{\text{boost}}), \quad k \in \{1,...,D_{\text{boost}}\}
\end{equation}

\subsubsection{Selective Critic Reset}
Rather than discarding all learned values, we apply a partial reset to both critics for all states when paired with information collection actions
\begin{equation}
    Q^{\bullet}(s,a) \leftarrow \lambda \cdot Q^{\bullet}(s,a), \quad \forall s \in \mathcal{S}, \forall a \in\mathcal{A}_{\text{collection}}
\end{equation}

where \(\lambda\ \in \mathbb{R}\) is the reset factor and \(\mathcal{A}_{\text{collection}}\) is specifically the set of information collection actions. This approach preserves previously learned knowledge while encouraging re-exploration of collection strategies throughout the entire state space after priority reordering.

\begin{table}[!t]
\renewcommand{\arraystretch}{1.2}
\caption{Performance Comparison (Single Priority Shift)}
\label{tab1:performance_comparison_single}
\centering
\resizebox{\linewidth}{!}{%
\begin{tabular}{@{}lccc@{}}
\toprule
\textbf{Metric} & \textbf{Standard Baseline} & \textbf{Boosted Baseline} & \textbf{CA-MIQ (Ours)} \\
\midrule
Mission Success (\%)    & 18.5   & 18.2    & \textbf{65.9} \\
Info Collection (\%)    & 20.8   & 20.4   & \textbf{69.2} \\
Recovery Success (\%)   & 0.0    & 0.0    & \textbf{100.0} \\
Recovery Time (Ep.)     & N/A    & N/A    & \textbf{685} \\
Reward (per Ep.)        & 4.2   & 3.9   & \textbf{63.9} \\
\bottomrule
\end{tabular}
} 
\vspace{0.5em}
\end{table}

\begin{table}[ht]
\renewcommand{\arraystretch}{1.2}
\caption{Performance Comparison (Multiple Priority Shifts)}
\label{tab2:performance_comparison_multi_shift}
\centering
\resizebox{\linewidth}{!}{%
\begin{tabular}{@{}lccc@{}}
\toprule
\textbf{Metric} & \textbf{Standard Baseline} & \textbf{Boosted Baseline} & \textbf{CA-MIQ (Ours)} \\
\midrule
Mission Success (\%)    & 12.3   & 13.1    & \textbf{50.2} \\
Info Collection (\%)    & 14.5   & 15.3   & \textbf{52.9} \\
Recovery Success (\%)   & 0.0    & 0.0    & \textbf{100.0} \\
Recovery Time (Ep.)     & N/A    & N/A    & \textbf{766} \\
Reward (per Ep.)        & -6.1   & -5.1   & \textbf{42.7} \\
\bottomrule
\end{tabular}
} 
\vspace{0.5em}
\end{table}

\begin{table}[ht]
\renewcommand{\arraystretch}{1.3}
\caption{Ablation Study Results (Multiple Priority Shifts)}
\label{tab3:ablation_study}
\centering
\resizebox{\linewidth}{!}{%
\begin{tabular}{@{}lccc@{}}
\toprule
\textbf{Configuration} & \textbf{Adapt. Time (Ep.)} & \textbf{Mission Success} & \textbf{Info Collection} \\
\midrule
Full CA-MIQ         & \textbf{766}  & \textbf{50.2\%}  & \textbf{52.9\%} \\
\midrule
w/o Priority Alignment + Awareness     & 808.7   & 47.3\%  & 50.3\%  \\
                           & (-5.6\%) \(\downarrow\) & (-5.8\%) \(\downarrow\) & (-4.9\%) \(\downarrow\) \\
                           \midrule
w/o State Novelty      & 851.2   & 34.4\%  & 43.8\%  \\
                           & (-11.1\%) \(\downarrow\) & (-31.5\%) \(\downarrow\) & (-17.2\%) \(\downarrow\) \\
\midrule
w/o Exploration Boost      & 780   & 49.2\%  & 52.2\%  \\
                           & (-1.9\%) \(\downarrow\) & (-2.0\%) \(\downarrow\) & (-1.4\%) \(\downarrow\) \\
\midrule
w/o Selective Reset        & 777.4   & 49.3\%  & 52.2\%  \\
                           & (-1.5\%) \(\downarrow\) & (-1.8\%) \(\downarrow\) & (-1.4\%) \(\downarrow\) \\
\midrule
Intrinsic Reset Only      & 770   & 49.5\%  & 52.7\%  \\
                           & (-0.5\%) \(\downarrow\) & (-1.4\%) \(\downarrow\) & (-1.3\%) \(\downarrow\) \\
\midrule
Extrinsic Reset Only      & 770.3   & 49.6\%  & 52.7\%  \\
                           & (-0.6\%) \(\downarrow\) & (-1.2\%) \(\downarrow\) & (-0.5\%) \(\downarrow\) \\
\bottomrule
\end{tabular}
} 
\caption*{\footnotesize\textit{Note:} Values in parentheses show the percent change relative to Full CA-MIQ; arrows indicate whether that change is an increase (\(\uparrow\)) or a decrease (\(\downarrow\)) in performance.}
\end{table}

\section{Experiments}
\subsection{Environment}
To evaluate our approach, we implemented a custom OpenAI Gym gridworld environment , similar to \cite{panagopoulos2024selective}, designed to reflect realistic SAR mission dynamics with changing information priorities. The environment features a $4 \times 4$ grid where an autonomous agent must collect mission-critical information types (denoted as $\mathtt{X}$, $\mathtt{Y}$, and $\mathtt{Z}$) in a specific priority order before reaching a designated target location to complete the mission. The state and action spaces follow the same structure as in \cite{panagopoulos2024selective}. No hierarchical agents are tested in this setting. A mission is considered successful when the agent collects all required information in the correct order, reaches the target, and saves the victim. The environment assigns rewards for movement, falling into ditches, exceeding action limits at information locations, successful information collection, and completing the mission. Additionally, it provides a method to dynamically swap the priority order of the information types at pre-defined episodes, requiring the agent to detect and adapt its policy without resetting the learned Q-values. The code is available at \url{https://github.com/dimipan/DynamicPriorities}.

\subsection{Baselines}
We implemented and compared the following agents to evaluate our approach:

\begin{enumerate}
    \item \textbf{Q-Learning (Baseline)} - A standard Q-learning agent using \(\varepsilon\)-greedy exploration with environment-based rewards only.

    \item \textbf{Q-Learning + $\varepsilon$-Boost} - Identical to 1) but temporarily boosts~$\varepsilon$ for $50$ (\(D_{\text{boost}}\)) episodes immediately after a priority change.

    \item \textbf{CA-MIQ (ours)} - Dual-critic agent with extrinsic and intrinsic Q-tables, where the intrinsic critic adds (i) state–action novelty, (ii) information-location awareness, and (iii) priority-alignment (Sec.~\ref{subsec:camiq}). After a priority swap, only entries relevant to information locations are reset and an adaptive $\varepsilon$-boost is applied.
\end{enumerate}

\subsection{Experimental Scenarios}
We evaluated our approach in three scenarios designed to test adaptation and robustness to dynamic mission demands:

\begin{itemize}
    \item \textbf{Static Priorities}: Standard information collection with fixed priorities (X→Y→Z throughout training), serving as a baseline test case.
    
    \item \textbf{Single Priority Shift}: One priority change (one update \(\Delta\pi\)) at episode 1700, where the collection order shifts from X→Y→Z to Y→Z→X, testing the agent's ability to adapt to a single contextual change.
    
    \item \textbf{Multiple Priority Shifts}: Two priority changes (two updates \(\Delta\pi\)), at episodes 1700 and 3500, with collection orders changing from X→Y→Z to Y→Z→X and then to Z→X→Y, testing the agent's ability to handle multiple contextual shifts over time.
\end{itemize}

Hyper-parameters are shared across all agents: learning-rate $\alpha=0.1$, discount factor $\gamma=0.99$, initial exploration rate \(\varepsilon_{0}=1.0\), minimum exploration rate \(\varepsilon_{\min}=0.1\), with linear decay over the full training horizon. For the CA-MIQ agent, the intrinsic reward is a weighted combination of novelty-based exploration (30\%), information location awareness (40\%), and priority alignment (30\%), boost factor \(\lambda_{\text{boost}}=2\), reset factor \(\lambda=0.5\). For each scenario, we ran 100 independent training sessions, each consisting of 5000 episodes. To ensure robustness, the agent is tested in randomly selected environments from a pool of five predefined configurations, each with different layouts for the target position, agent starting location, ditches, and information locations. All environments maintain the $4\times4$ grid structure and require collecting three pieces of information in the correct sequence. Performance metrics were averaged across these runs to ensure statistical significance. Key metrics included mission success rate, information collection success rate (i.e., the agent solved the task of collecting the necessary types of information), recovery success and time after priority changes (latter is measured in episodes).

\begin{figure}[!t]
\centering
\begin{subfigure}[t]{0.5\textwidth}
    \includegraphics[width=\textwidth]{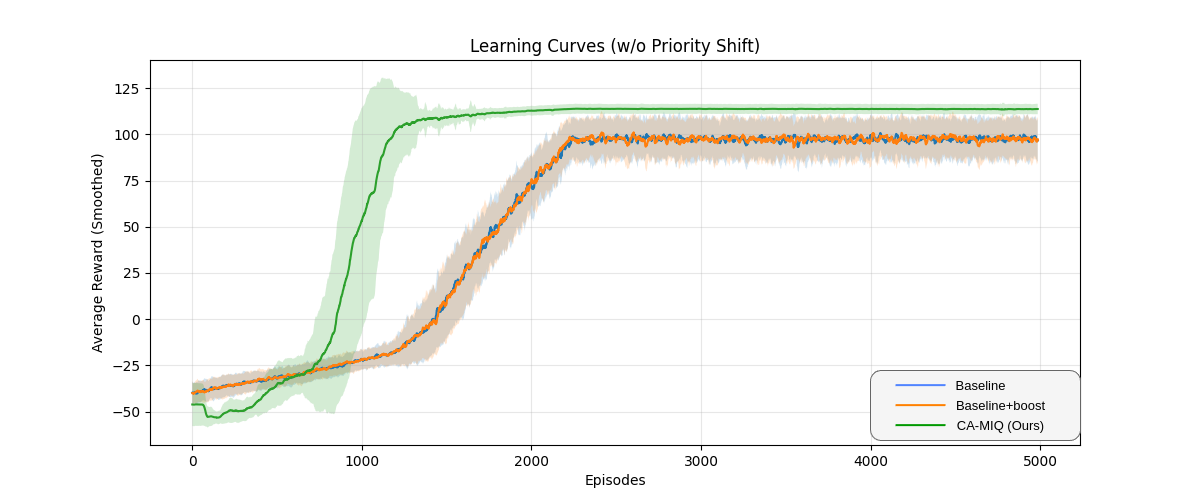}
    \caption{\textit{Static Priorities}}
    \label{fig:subfig1}
\end{subfigure}
\hfill
\begin{subfigure}[t]{0.5\textwidth}
    \includegraphics[width=\textwidth]{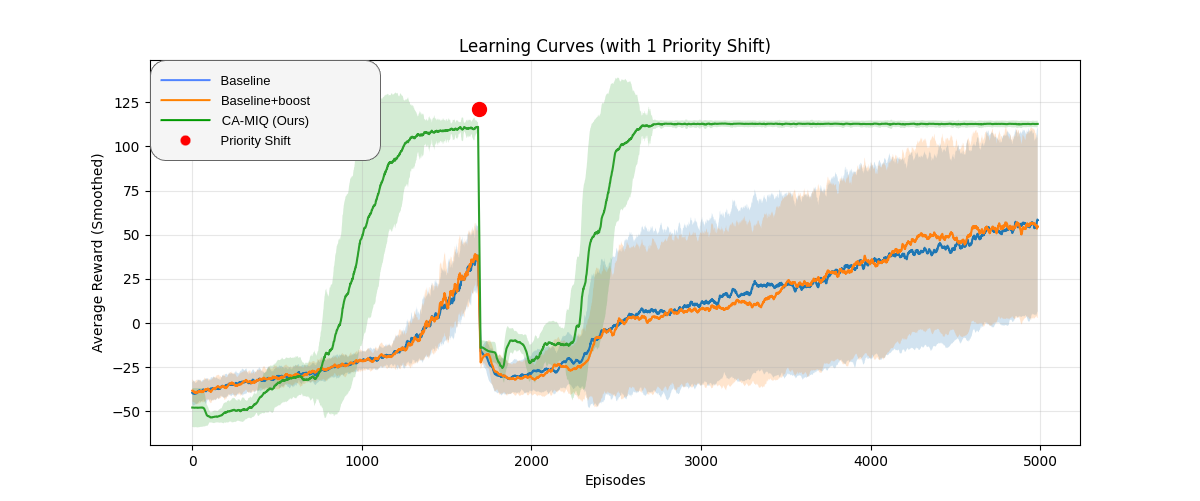}
    \caption{\textit{Single Priority Shift}}
    \label{fig:subfig2}
\end{subfigure}
\hfill
\begin{subfigure}[t]{0.5\textwidth}
    \includegraphics[width=\textwidth]{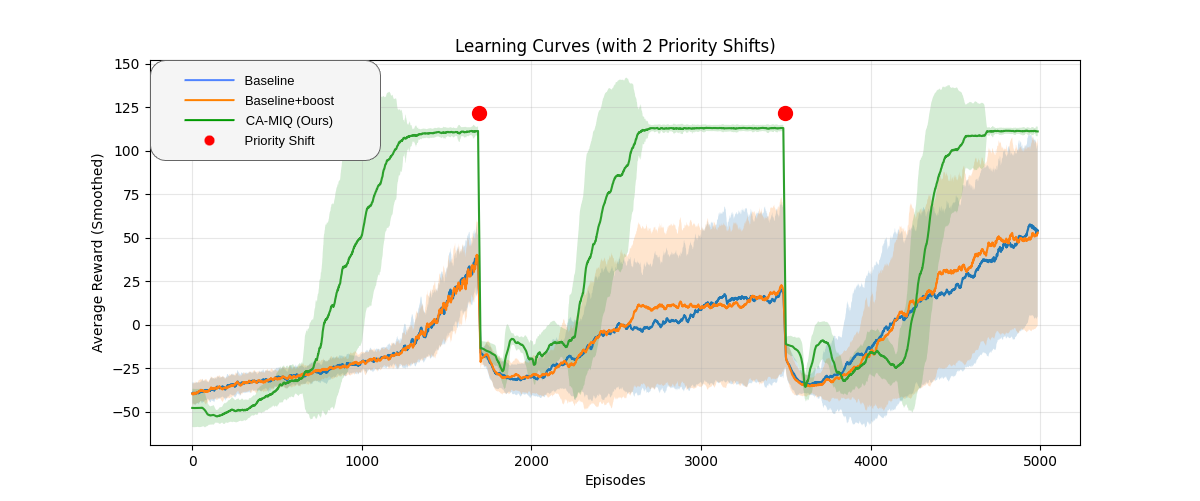}
    \caption{\textit{Multiple Priority Shifts}}
    \label{fig:subfig3}
\end{subfigure}
\caption{Comparison of reward trends under static and dynamic conditions.}
\label{fig:combined}
\end{figure}

\section{Results \& Discussion}
\subsection{Overall Performance Comparison}
The experimental results demonstrate that CA-MIQ outperforms both baseline with and without boosting across all metrics. As shown in Table~\ref{tab1:performance_comparison_single}, CA-MIQ achieved a mission success rate of 65.9\% in the single priority shift scenario, more than three times higher than both standard baseline (18.5\%) and boosted baseline (18.2\%). Similarly, information collection success rates were substantially higher with CA-MIQ (69.2\%) compared to the baselines (20.8\% and 20.4\%, respectively). The performance gap is also notable in the more challenging multiple priority shifts scenario (Table~\ref{tab2:performance_comparison_multi_shift}), where CA-MIQ maintained a 50.2\% mission success rate while both baseline approaches struggled with success rates below 15\%. This demonstrates CA-MIQ's robustness when faced with multiple contextual changes, which better reflects the dynamic nature of real-world SAR operations.

\subsection{Priority Adaptation Mechanisms}
CA-MIQ's performance stems from two key mechanisms. First, transient epsilon boosting temporarily increases exploration after detecting a priority shift, helping the agent escape local optima established under previous priority orderings. Second, selective critic reset preserves general environment knowledge while targeting only value estimates related to information collection for relearning. This balanced approach allows CA-MIQ to discover new optimal policies efficiently while maintaining valuable knowledge about the environment's dynamics, explaining its sustained performance even after multiple priority shifts. Results show that CA-MIQ achieved a 100\% recovery success rate in both single shift and multiple shifts scenarios, whereas both baseline approaches completely failed to adapt (0\% recovery success). The average recovery time for CA-MIQ was 685 episodes after a single shift and 766 episodes for multiple shifts, showing consistent adaptation capabilities even in complex scenarios.

\subsection{Learning Dynamics and Stability}
The learning curves in Figure~\ref{fig:combined} provide further insights into the behavior of each approach. In the static environment (Figure~\ref{fig:subfig1}), all three approaches eventually converge to similar performance levels, though CA-MIQ learns more efficiently, reaching higher rewards earlier in training. In the single priority shift scenario (see Figure~\ref{fig:subfig2}), immediately after the priority change (marked by a red dot), all approaches experience a performance drop as expected. While the baseline methods struggle to recover, with their performance remaining below pre-shift levels, CA-MIQ rapidly adapts and returns to near-optimal performance levels. The multiple priority shifts scenario (Figure~\ref{fig:subfig3}) further emphasises this pattern. After each shifts, CA-MIQ experiences a temporary performance drop but consistently recovers successfully. The baseline approaches show minimal recovery after shifts, demonstrating CA-MIQ's stability and resilience to repeated contextual changes.

\subsection{Ablation}
The ablation study (Table~\ref{tab3:ablation_study}) provides empirical validation of each component's contribution to overall CA-MIQ's performance. The most critical finding is that removing state novelty has the most severe impact, reducing mission success by 31.5\% and information collection by 17.2\%. This suggests that novelty-driven exploration is crucial for achieving long-term mission goals in dynamic environments. Removing priority alignment and awareness components resulted in a 5.8\% decrease in mission success and 4.9\% decrease in information collection, confirming that guiding exploration based on current priorities is essential for effective adaptation. Interestingly, while the exploration boost mechanism improved mission success by 2.0\% and information collection by 1.4\%, its impact was less dramatic than the novelty and priority alignment components. This suggests that the core intrinsic motivation components are more fundamental to CA-MIQ's success than the boost mechanism.

\subsection{Limitations}
The effectiveness of our selective reset mechanism depends on the structure of the SAR task and the nature of the encountered priority shifts. Currently, adaptation following a priority shift requires approximately 600–800 episodes, a number which might be reduced through adaptive reset scheduling or parameterised exploration boosts. Additionally, the fixed coefficients balancing novelty, information-location awareness, and priority alignment are effective in our SAR scenarios but lack generalisability. An adaptive weighting mechanism could automatically adjust these parameters based on mission characteristics and agent performance. System robustness depends on accurate and timely detection of context changes. Incorporating Bayesian-based detection or distribution-shift monitoring would improve reliability without requiring explicit signals. Furthermore, while our implementation avoids the computational overhead inherent to the ensemble-based uncertainty estimation employed in \cite{sukhija2024maxinforl}, it is inherently limited to discrete state–action spaces. Although we deliberately chose a simplified grid environment to isolate and clearly demonstrate our algorithm's performance at the high-level decision-making scale relevant to SAR operations, validating it in partially observable domains remains essential future work.

\section{Conclusion}
We introduce CA‑MIQ (Context‑Aware Max‑Information Q‑learning) to address a critical challenge in RL-based systems: adapting to dynamic changes in the priority order of information types the agent is expected to focus on. The frameowrk enhances information-gathering strategies when mission objectives shift unexpectedly by integrating dual critics to encourage the reduction of priority-driven uncertainty with targeted exploration boosting and selective reset mechanisms. Our approach demonstrates up to 4× higher mission success rates compared to baselines and near-perfect recovery following priority shifts. The framework's strong performance in SAR-inspired simulations highlights its potential for broader application in domains requiring adaptive information gathering under evolving constraints. CA-MIQ effectively bridges the gap between algorithmic and priority-driven exploration, marking an important advancement for autonomous systems in dynamic, mission-critical environments.




\addtolength{\textheight}{-9.33cm}   


\bibliography{thebibliography}
\bibliographystyle{IEEEtran}

\end{document}